\documentclass[runningheads]{llncs}
\usepackage[T1]{fontenc}
\usepackage{graphicx}
\usepackage{subcaption}
\usepackage{float}
\usepackage{hyperref}
\usepackage{booktabs}
\usepackage{multirow}
\usepackage{pgfplots}
\pgfplotsset{compat=1.18}
\pgfplotsset{plot coordinates/math parser=false}
\begin{document}
\title{Single-agent vs. Multi-agents for Automated Video Analysis of On-Screen Collaborative Learning Behaviors}
\titlerunning{Single vs. Multi-Agent Systems for Learning Video Analytics}
\author{Likai Peng\inst{1}\orcidID{0000-0002-0475-0541} \and
Shihui Feng\inst{1,2}\orcidID{0000-0002-5572-276X}}
\authorrunning{L. Peng \and S. Feng}
%
\institute{Faculty of Education, The University of Hong Kong, Pokfulam, Hong Kong \\
\email{likaipen@connect.hku.hk} \and
Institute of Data Science, The University of Hong Kong, Pokfulam, Hong Kong\\
\email{shihuife@hku.hk}}
%
%
%
\maketitle              
\begin{abstract}
On-screen learning behavior provides valuable insights into how students seek, use, and create information during learning. Analyzing on-screen behavioral engagement is essential for capturing students’ cognitive and collaborative processes. The recent development of Vision Language Models (VLMs) offers new opportunities to automate the labor-intensive manual coding often required for multimodal video data analysis. In this study, we compared the performance of both leading closed-source VLMs (Claude-3.7-Sonnet, GPT-4.1) and open-source VLM (Qwen2.5-VL-72B) in single- and multi-agent settings for automated coding of screen recordings in collaborative learning contexts based on the ICAP framework. In particular,  we proposed and compared two multi-agent frameworks: 1) a three-agent workflow multi-agent system (MAS) that segments screen videos by scene and detects on-screen behaviors using cursor-informed VLM prompting with evidence-based verification; 2) an autonomous-decision MAS inspired by ReAct that iteratively interleaves reasoning, tool-like operations (segmentation/ classification/ validation), and observation-driven self-correction to produce interpretable on-screen behavior labels. Experimental results demonstrated that the two proposed MAS frameworks achieved viable performance, outperforming the single VLMs in scene and action detection tasks. It is worth noting that the workflow-based agent achieved best on scene detection, and the autonomous-decision MAS achieved best on action detection. This study demonstrates the effectiveness of VLM-based Multi-agent System for video analysis and contributes a scalable framework for multimodal data analytics. 

\keywords{Multi-agent System  \and Behavior Coding \and Multimodal Learning Analytics.}
\end{abstract}
%
%
\section{Introduction}
Understanding students’ behavioral engagement is critical for understanding how learning occurs and for developing effective teaching and intervention strategies\cite{gomes2023modeling}. To capture, analyze and visualize this engagement, researchers increasingly rely on multimodal learning process data, which integrates diverse sources such as interaction logs, oral recordings, written outputs, eye tracking data, spatial movements, and facial expressions\cite{schiller2024understanding,feng2024heterogenous}. Although multimodal data has been widely used to analyze engagement, screen recordings, capturing both on-screen actions and contextual scenes, remain underexplored\cite{sharma2020multimodal}. Screen recordings offer distinct advantages over traditional data sources\cite{sundar2014user}. Unlike log data, which captures discrete events (e.g., clicks or page visits), screen recording data show more detailed and fine-grained records of on-screen actions and learning context. However, current analyses of screen recordings heavily rely on manual coding, which becomes impractical for large datasets or complex coding schemes (e.g., multidimensional collaboration behaviors)\cite{erkens2008automatic}. Recent advances in computer vision, exemplified by the Computer Vision for Position Estimation (CVPE) framework\cite{li2023cvpe}, have demonstrated the feasibility of scalable, privacy-preserving extraction of socio-spatial behaviors from video data. However, automated approaches to on-screen action recognition in educational contexts are still in their early stages of development. Vision language model (VLM) has demonstrated a high level of zero-shot performance in a variety of visual tasks including action recognition and understanding\cite{bosetti2024text}. While VLM offers new possibilities for understanding onscreen behavior, it still lags behind in fine-grained behavior recognition and temporal understanding\cite{wang2024grounded}. 

Multi-agent collaboration can improve video understanding by enabling complementary specialists (e.g., grounding and vision agents) to gather evidence and support multi-step reasoning, outperforming non-agent baselines\cite{liu2025longvideoagent}. This study aims to assess different agent systems using VLMs in on-screen behavior analysis, including workflow-based and autonomous decision-making agent systems. We develop a new method leveraging VLMs and multi-agent techniques to automatically analyze learning behavior using on-screen data based on the ICAP framework\cite{chi2014icap}. Existing AI in education research has predominantly centered on LLMs, by integrating LLMs into the learning process (e.g., AI-supported peer feedback) or assessment\cite{gao2023coaicoder}. However, the use of VLMs holds considerable potential to enhance understanding of multimodal data in educational contexts, as they can facilitate deeper understanding of learning processes by integrating visual and textual information for tasks such as image captioning, visual question answering, and visual search \cite{zhang2024vision}. This study explored the following research questions. 

\textbf{RQ1:} To what extent can single VLMs (commercial: GPT-4.1, Claude-3.7-Sonnet; open-source: Qwen2.5-VL-72B) accurately code onscreen behaviors from video in a few-shot setting? 

\textbf{RQ2:} What was the performance of multi-agent orchestration strategies (workflow-based vs. autonomous decision-making) on coding onscreen behaviors compared to single-agent baselines? 

\section{Related Work}
\subsection{On-Screen Learning Behavior}
The ICAP framework categorizes learner engagement into Passive, Active, Constructive and Interactive modes offering a theoretical lens for interpreting observable behaviors as cognitive engagement\cite{chi2014icap}. Applied across diverse contexts from video-based STEM activity recognition to robot-supported collaborative learning—ICAP has proven effective for analyzing on-screen behaviors and inferring engagement level\cite{lee2023exploring,castro2019interactive}. 

Prior work has examined on-screen behaviors at both individual and group levels. Sundar compared six different on-screen actions (click-to-download, drag, mouseover, slide, zoom, and 3D carousel) and found that certain information-based actions, such as mouseover, provided a more positive user experience compared to others like zoom, while some navigation actions, such as slide, appeared to be more effective than alternatives like drag\cite{sundar2014user}. In online learning environments, Cheung used verbal exchanges and on-screen interactions among three pairs of learners to explore collaborative writing patterns and how they leveraged semiotic resources and multimodal elements, identifying that “Text type and revision” and “Internet search” were the most frequent on-screen activities in the collaborative writing process\cite{cheung2022verbal}. Oh and Sundar further linked interaction frequency (clicks, drags) to learners’ perceptions of online content\cite{oh2020happens}.

Despite these advances, existing studies mainly analyzed on-screen behavior within a constrained learning platform. Few prior studies have investigated students’ on-screen actions across platforms within a learning task, such as generative AI applications, web search engines, and shared group documents. This is critical for understanding how students navigate and integrate diverse digital resources in authentic, multi-tool learning environments.

\subsection{Generative AI-assisted Video Analysis}
\subsubsection{VLMs for Video Analysis}
VLMs integrate computer vision and natural language processing to interpret multimodal content, demonstrating strong potential and capabilities in video analytics, including multimodal emotion recognition and collaborative activities\cite{teotia2024evaluating}. Gao et al. proposed T3-Agent for tuning multi-modal agents, which automatically synthesizes tool-usage data and finetunes VLMs as controllers for sophisticated multi-modal reasoning which improved VLM reasoning capabilities by 20\% through automated tool-usage data \cite{gao2025multimodalagenttuningbuilding}.

In educational applications, VLMs have been used to extract and interpret images and videos. Alahmadi and Alshangiti demonstrated that LLMs such as GPT-4V and Gemini outperform traditional optical character recognition (OCR) engines in transcribing source code from screencasts across various video qualities. These results indicate the robustness and advanced capabilities of VLMs in diverse educational scenarios\cite{alahmadi2024optimizing}. However, significant performance gaps remain—AutoEval-Video benchmarks reveal VLMs achieve only 32.2\% accuracy compared to 72.8\% human performance in video question answering, particularly struggling with temporal and dynamic comprehension critical for educational contexts\cite{chen2024autoeval}.

\subsubsection{Multi-agent System (MAS) for Video Analysis}
Multi-agent systems mitigate VLM's limitations through distributed coordination \cite{falco2020tendencies}. Educational applications leverage agent specialization: Jiang et al.’s von Neumann framework decomposes curriculum design into modular tasks (knowledge graphs, resource matching, difficulty calibration) processed by specialized agents\cite{jiang2024aiagenteducationvon}. This close collaboration enables the creation of adaptive, dynamic course plans that are tailored to individual learning needs. Yang et al.’s Embodied Multi-Modal Agent (EMMA) integrates visual perception with LLM planning to generate efficient action sequences from pixel observations\cite{Yang_2024_CVPR}. Their Embodied Multi-Modal Agent (EMMA) addresses this challenge by training a VLM to align with visual world dynamics and distilling the skills of an LLM agent in a parallel text world, enabling the agent to take textual task instructions and pixel observations to produce efficient action sequences. Shriram et al. deployed MAS-VLM for zero-shot hazardous object detection in traffic videos, demonstrating semantic event analysis beyond static prediction\cite{shriram2025towards}. 

Despite the proven capabilities of VLMs in open-ended video QA tasks and visual retrieval\cite{chen2024autoeval,shriram2025towards}, there is currently little research that combines VLM-based MAS frameworks with pedagogical theories to analyze student behavior in collaborative learning settings. 

\section{Methods}
\subsection{Research Design Overview}
We adopt a comparative experimental design to address two research questions across two parts: (1) few-shot VLM prompting in VLMs (RQ1) and (2) the workflow-based MAS and ReAct-style MAS (RQ2). To address RQ1, we designed few-shot prompts to to evaluate the classification accuracy of VLMs in two aspects, including basic and complex questions, 1) what on-screen scenarios were included in given video, and 2) what actions were detected in it. Regarding the design of MAS, we compared the performance of both workflow-based MAS and ReAct-style MAS on on-screen behavior detection tasks. 

\subsection{The Learning Context}
The present study involved 27 undergraduate students from a research university, who were organized into nine groups of three. In a laboratory setting, each group participated in a 30-minute collaborative problem-solving (CPS) activity designed to support learning of Bloom's taxonomy or ICAP framework. The task demanded that students interpret the target theory collaboratively. Initially, the participants were invited to engage in a group discussion on the theoretical definition. This was followed by the administration of a series of multiple-choice questions and the composition of a concise group essay that synthesized their comprehension of the subject.

Each student was equipped with their own personal laptop, and group members collaborated through a combination of discussion, information seeking, and shared writing. During the task, students moved across multiple digital environments, including Generative-AI (GAI) tools, web search engines, and shared group documents. The design of the system resulted in the creation of a screen-based collaborative workspace.

\begin{figure}[htbp]
    \centering
    \subfloat[GAI Interface]{
        \includegraphics[width=0.30\textwidth,height=3.2cm,keepaspectratio]{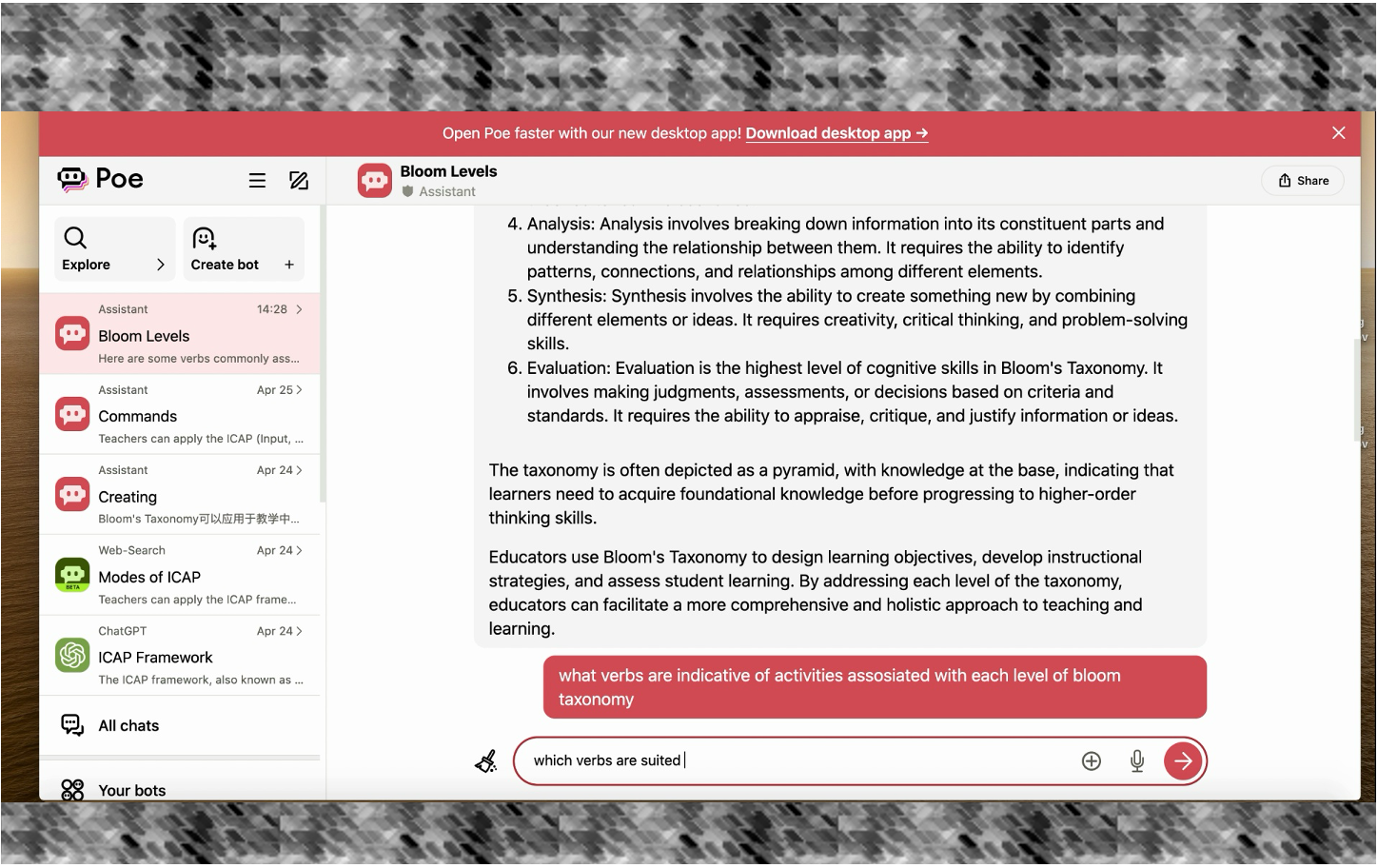}
    }
    \hfill 
    \subfloat[Web Content]{
        \includegraphics[width=0.30\textwidth,height=3.2cm,keepaspectratio]{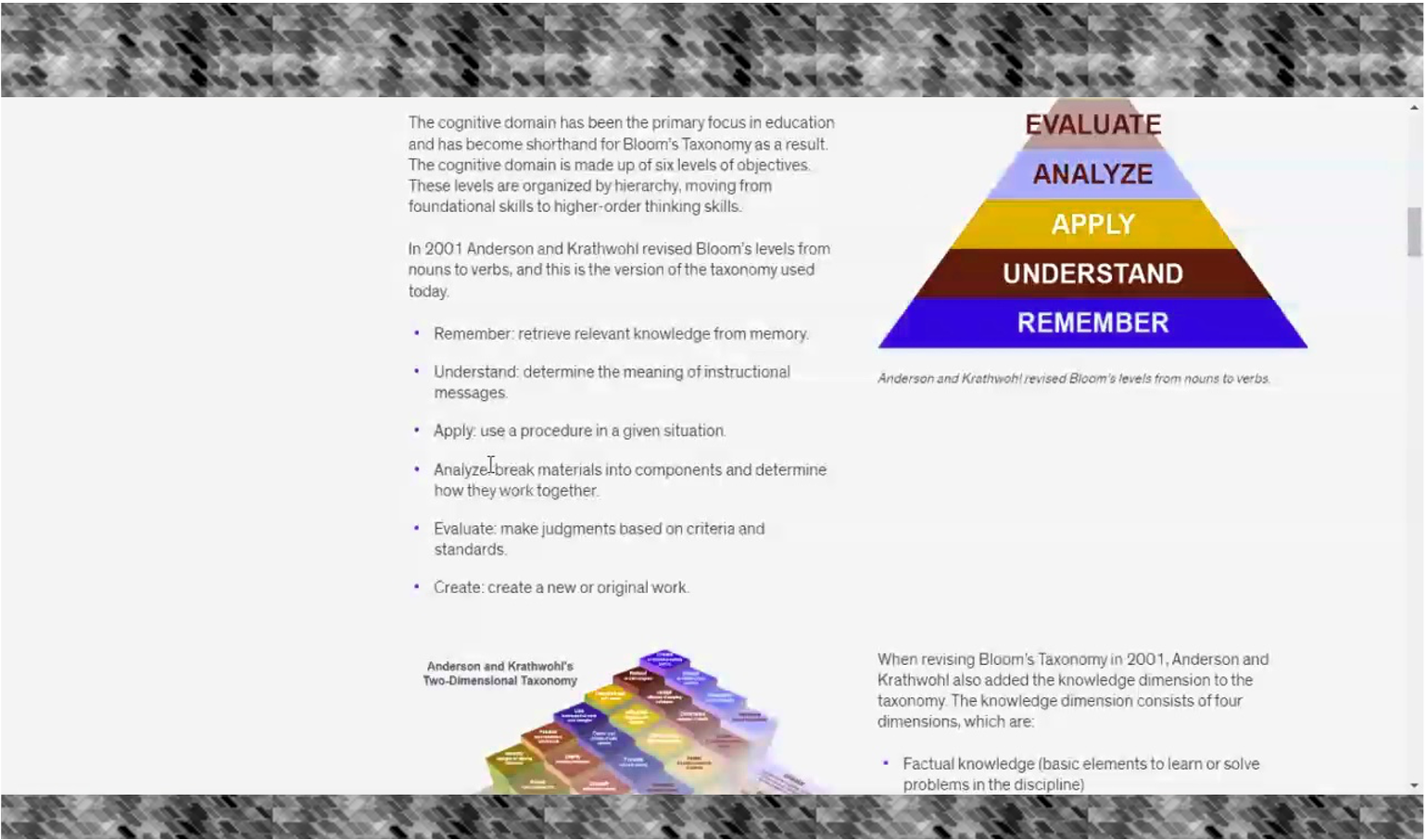}
    }   
    \hfill
    \subfloat[Group Document]{
        \includegraphics[width=0.30\textwidth,height=3.2cm,keepaspectratio]{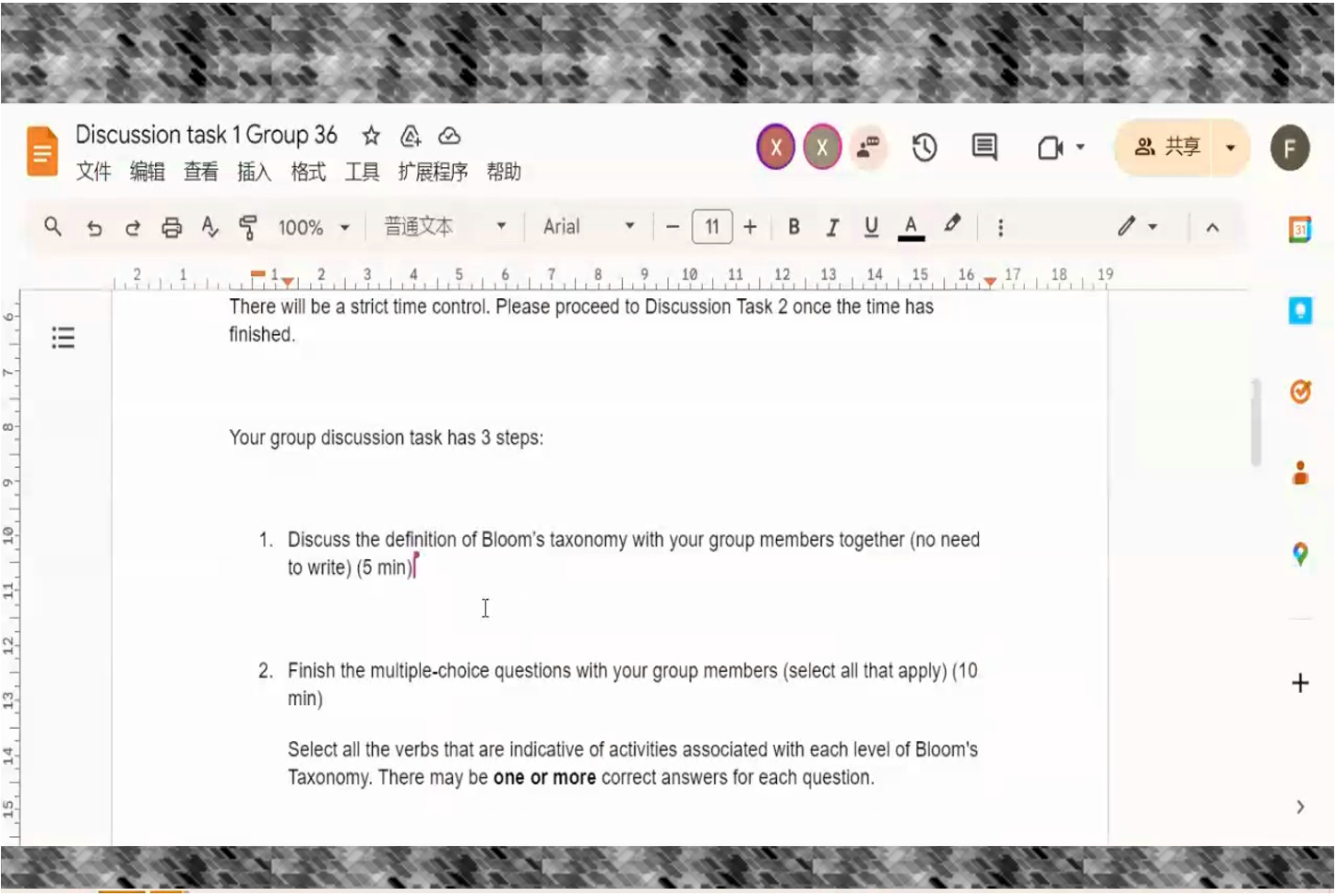}
    }   
    \caption{Screenshot Data Examples}
    \label{fig:screenshot_data_example}
\end{figure}

The present study's learning context proved to be a particularly suitable environment for analysis due to the observable traces of collaborative learning processes it provided. It is evident that the screen recordings captured not only the digital tools being utilized, but also the manner in which students coordinated their activity across disparate interfaces while engaged in a shared collaborative learning task. Figure \ref{fig:screenshot_data_example} shows three examples of video screenshots from different scenes.

\subsection{Data Collection}
\subsubsection{Screen Data Collection}
Screen recordings were captured via Zoom to document students’ on-screen behaviors during collaborative tasks. The recordings captured typing, clicking, scrolling, page navigation, and contextual information about platforms (e.g., GAI interfaces, web search engines, group documents).

\subsubsection{Coding Framework}
We operationalized the ICAP framework\cite{chi2014icap} to code on-screen behaviors by decomposing them into two components: \textbf{Scenes}, defined as distinct on-screen contexts where learning occurs (generative AI interfaces, web search engines, shared group documents); and \textbf{Actions}, representing specific user interactions within scenes (co-editing, scrolling, reading with highlighting, copying text)

Each action is mapped to one of four ICAP engagement levels\cite{chi2014icap}. Passive engagement involves viewing information without active involvement (e.g., reading with scrolling). Active engagement encompasses interactions with materials without generating new knowledge, such as web searching or selecting answers. Constructive engagement involves producing ideas beyond given information, exemplified by prompting generative AI tools. Interactive engagement represents the co-construction of knowledge through collaboration, such as collaborative document editing. Additionally, screen freezing—where no detectable action occurs—is categorized as concealed engagement. This two-component framework enables detailed analysis of how students leverage digital resources during collaborative learning. The complete coding scheme is presented in Table \ref{tab:coding_framework}.
\begin{table}[H]
  \renewcommand{\arraystretch}{1.22}
  \centering
  \caption{Operationalization of ICAP Framework for On-Screen Behaviors}
  \label{tab:coding_framework}
  \small
  \setlength{\tabcolsep}{6pt}
  \begin{tabular}{p{1.6cm} p{1.2cm} p{4.3cm} p{5cm}}
    \toprule
    \textbf{Dim.} & \textbf{Scene} & \textbf{Action} & \textbf{Description} \\
    \midrule

    \multirow{4}{*}{Active}
      & Web
      & Searching Internet
      & Searching information using search engines or other digital sources. \\
    & Docs
      & Ticking Answers
      & Completing or correcting answers in multiple-choice questions. \\
    & All
      & Reading with Highlighting
      & Reading while highlighting text on screen using the cursor/mouse. \\
    & All
      & Copy and Paste
      & Copying selected content and pasting it into the shared group document. \\
    \addlinespace[2pt]

    Constructive
      & GAI
      & Prompting GAI
      & Constructing prompts for a generative AI tool. \\
    \addlinespace[2pt]

    Interactive
      & Docs
      & Group Document Co-Editing
      & Editing a shared group document, including deleting, adding, and revising content. \\
    \addlinespace[2pt]

    Passive
      & All
      & Reading with Scrolling
      & Reading content while scrolling up/down (optionally with cursor movement). \\
    \addlinespace[2pt]

    Conceal
      & All
      & Freezing
      & No on-screen actions detected. \\
    \bottomrule
  \end{tabular}

  \vspace{2pt}
  \footnotesize{\textit{Note.} Scenes: GAI = Generative AI interface; Web = Web content; Docs = Group documents; All = \{GAI, Web, Docs\}.}
\end{table}

\subsection{Data Preprocessing }
\subsubsection{Video Segmentation}
To adapt to VLM's input limitations, we segment raw screen recordings into fixed 20-second sub-videos and sample 1 frame per second as VLM's input. We sampled and labelled 507 videos for evaluation. The 20-second segments were used as the evaluation units in this study. 

\subsection{Experiments}
\subsubsection{\textbf{Few-shot Single-agent Evaluation (RQ1)}}
To establish performance baselines for ICAP-informed on-screen behavior classification, we evaluated three leading VLMs in few-shot settings across the 507 held-out test videos.

\paragraph{\textbf{Model Selection}}
Model selection balanced scale, accessibility, and multimodal reasoning capability. The open-source model (\textbf{Qwen2.5-VL-72B}) represents the current parameter efficiency frontier in video understanding, while proprietary models (\textbf{GPT-4.1}, \textbf{Claude-3.7-Sonnet}) serve as performance ceilings due to their superior benchmark results on multimodal reasoning tasks \cite{bai2025qwen25vltechnicalreport,openai2025gpt41,anthropic2025claude37}. 
All selected models support extended video context (more than 20s at 1 fps sampling), ensuring methodological consistency across commercial and open-weight architectures. 

\paragraph{\textbf{Prompt Design}}
We evaluated a few-shot prompting strategies to assess the capability of models with contextual information. Full templates could be found at \href{https://anonymous.4open.science/r/AIED2026_submission_Appendix-D482/Appendices.md}{appendix C}. 

\subsubsection{\textbf{Multi-agent Workflow System (RQ2)}}
\paragraph{\textbf{The workflow-based MAS}}
The proposed workflow-based MAS implements a comprehensive approach for analyzing screen recordings through three collaborative components: Scene Detection and Segmentation (OSDS), Interaction-Aware Contextual Visual Prompting (ICVP), and Evidence-Validated Behavior Mapping (EVBM). The architecture balances computational cost and efficiency by employing VLMs for high-level semantic reasoning (scene and action understanding) and traditional computer vision algorithms for efficient cursor tracking. This hybrid approach leverages the complementary strengths of each technique: VLMs handle semantic tasks, while CV algorithms address low-level visual pattern matching. The three parts of the MAS system are presented in Figure \ref{fig:system arch} and illustrated below. 

\begin{figure}[h]
  \centering
  \includegraphics[width=\linewidth]{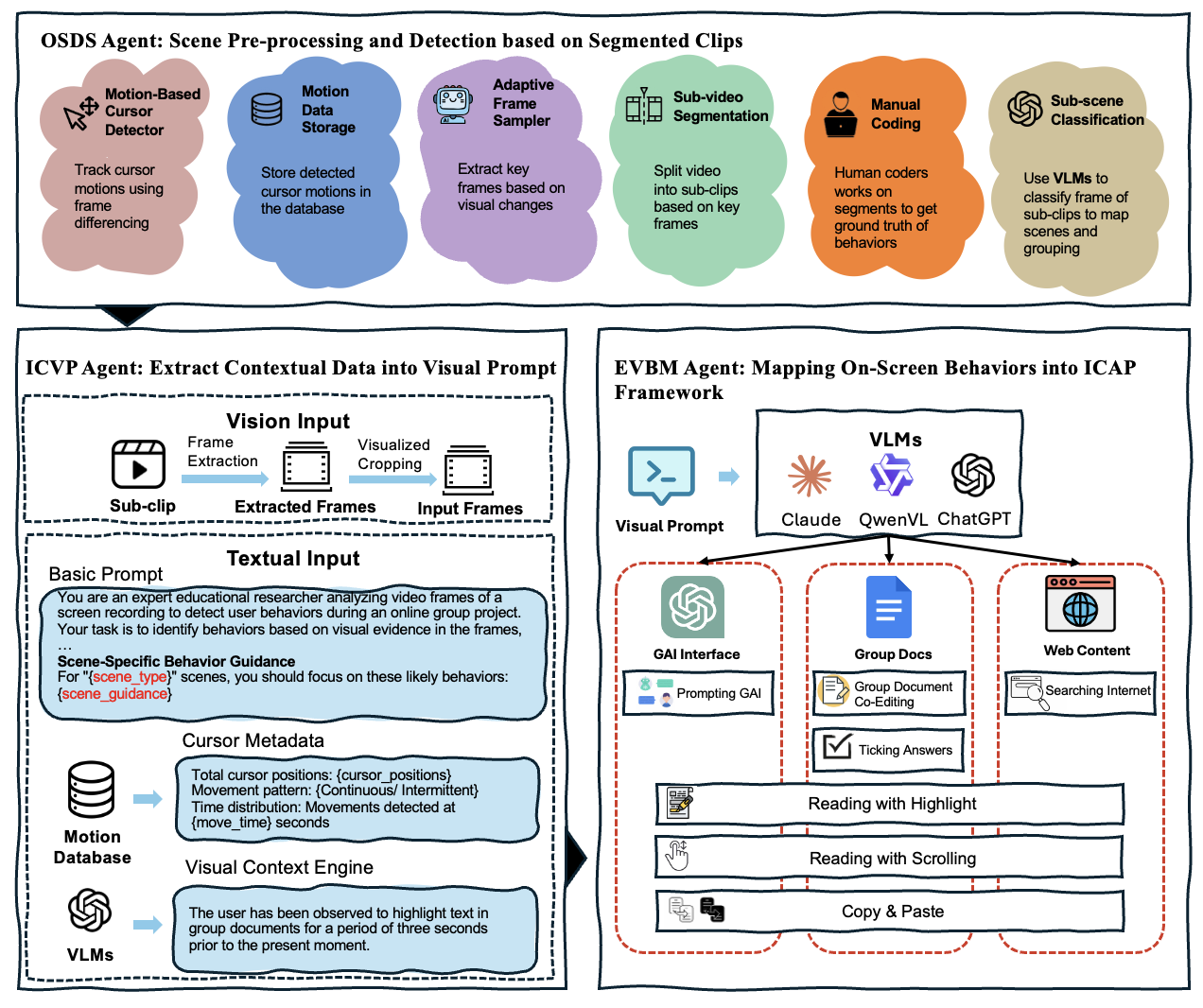}
  \caption{The System Architecture of Multi-agent for Video Analysis of On-screen Behavior Detection}
  \label{fig:system arch}
\end{figure}

\paragraph{\textbf{OSDS: Scene Detection and Segmentation}}
The OSDS agent segments videos into scene-specific sub-videos with classified scene types. First, it employs adaptive frame sampling based on visual changes to identify keyframes, reducing redundant processing while maintaining semantic coverage. These keyframes enable video segmentation into coherent content blocks. Second, a vision-language model classifies the initial frame of each sub-clip (Web Content, GAI Interface, or Group Documents). Third, adjacent sub-clips of the same scene type are aggregated, yielding scene-homogeneous segments. This approach significantly reduces computational overhead compared to analyzing full video data. Scene detection prompts are available at \href{https://anonymous.4open.science/r/AIED2026_submission_Appendix-D482/Appendices.md}{appendix A}

\paragraph{\textbf{ICVP: Interaction-Aware Contextual Visual Prompting}}
The ICVP agent enhances behavior detection by integrating cursor information with scene-specific guidance. \textit{Visual input} consists of sub-video frames extracted at regular intervals. A visual highlight (bounding box) is overlaid to direct attention to regions of interest without obscuring details. \textit{Textual input} comprises three components: (1) cursor trajectories from frame differencing, including position, movement pattern, and temporal distribution; (2) scene-specific contextual guidance that delineates explicit detection criteria for each behavior type within that scene; (3) a task description specifying behavior definitions, output format, and task objectives. This integration of cursor dynamics with scene. The prompts are available at \href{https://anonymous.4open.science/r/AIED2026_submission_Appendix-D482/Appendices.md}{appendix B}.

\paragraph{\textbf{EVBM: Evidence-Validated Behavior Mapping}}
The EVBM agent validates detected behaviors through evidence-based filtering. Rather than applying arbitrary confidence thresholds, the system correlates cursor movement patterns with detected visual evidence and applies feature-specific validation criteria. For example, ``Reading with Scrolling'' is validated by detecting both content movement and cursor trajectories consistent with reading behavior. This multi-modal evidence integration significantly enhances detection accuracy and reduces false positives while maintaining interpretability within educational video contexts.

\paragraph{\textbf{Workflow Integration}}
Following this three-agent process, each video segment is characterized by: (1) scene type, (2) detected behaviors with confidence scores, (3) supporting cursor trajectories, and (4) visual evidence explanations. The explicit reasoning traces from each agent support the educator's interpretability.

\subsubsection{\textbf{ReAct-style Multi-agent System}}
\begin{figure}[h]
  \centering
  \includegraphics[width=\linewidth]{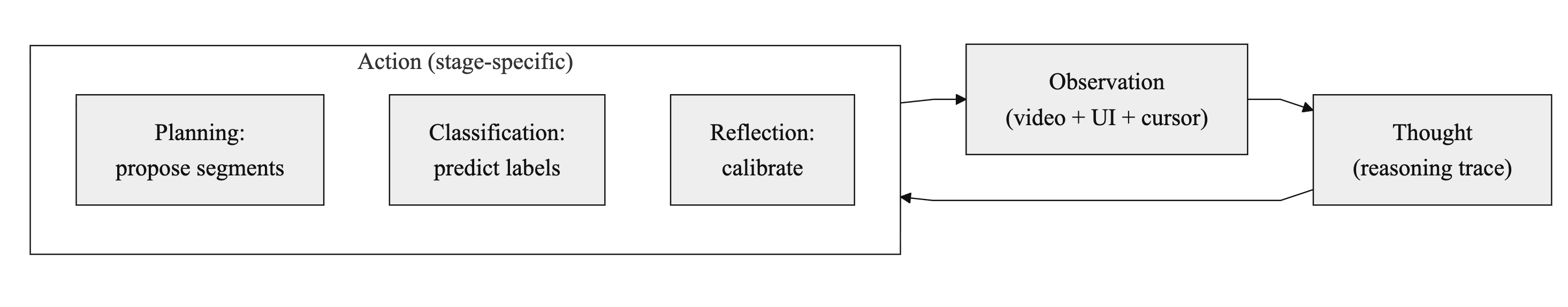}
  \caption{The System Architecture of ReAct-style Multi-agent System for Video Analysis of On-screen Behavior Detection}
  \label{fig:react-mas}
\end{figure}

Our system is inspired by ReAct, which interleaves reasoning traces and task-specific actions, where actions interface with external sources (e.g., tools or environments) to gather additional information and reduce error propagation~\cite{yao2022react}(see \ref{fig:react-mas}). Following the definition of autonomous agents as systems that operate in an environment, sense it, and act over time to pursue goals~\cite{wang2024survey}, we model on-screen behavior analysis as an iterative evidence-seeking process. Our system has the following characteristics. \textit{First, explicit reasoning traces}: the system generates detailed reasoning explanations at Planning, Classification, and Reflection stages, clarifying the rationale for each decision. These reasoning traces enhance interpretability. \textit{Second, interleaved reasoning and acting}: unlike linear workflow execution, our three-stage architecture realizes reasoning$\rightarrow$action$\rightarrow$observation$\rightarrow$adjustment cycles. Planning stages reason about segmentation points and act by outputting segments; Classification stages reason about behavior labels and act by producing predictions; Reflection stages reason about confidence calibration and act by flagging anomalies. \textit{Third, adaptive reasoning facing contradictory evidence}: when evidence from cursor, frame changes, and UI context contradict each other, the VLM reasons about the credibility of each evidence source. For instance, a static cursor may conflict with page content changes; the system reasons whether this represents ``Reading with Scrolling'' or ``Freezing.'' \textit{Fourth, self-correction mechanisms}: the Reflection stage checks scene-action compatibility, automatically lowering confidence for incompatible predictions and flagging them for review---a capability beyond workflow systems that typically fail when contradictions arise.

ReAct paradigm addresses three critical challenges in educational video analysis\cite{yao2022react}. First, real classroom student behaviors exhibit long-tail and diverse characteristics; exhaustive rules cannot cover all cases, whereas reasoning enables the system to generalize to novel behavior patterns. Second, multimodal information (cursor, frame changes, UI) in educational videos often contradicts or remains ambiguous; reasoning rather than rule-matching can address these uncertainties. Third, educational scenarios demand explainable decisions; ReAct's explicit reasoning traces satisfy educators' needs for transparent decision processes. Additionally, the ReAct architecture supports a more flexible experimental design — allowing us to independently test the contributions of the Planning, Classification, and Reflection stages.

\subsection{\textbf{Evaluation Metrics}}
To evaluate the VLM-performance, 507 sampled videos were manually labeled and checked. These labels serve as a benchmark for both VLM auto-coding evaluation scores. Two human experts independently coded the same segments to provide a reference standard. The inter-rater reliability (IRR) between the two coders was calculated using Cohen's Kappa, yielding a substantial agreement score of 0.945. Any discrepancies between the coders were discussed and resolved through consensus meetings, ensuring that the final ground truth labels accurately reflect shared expert judgment. This rigorous process strengthens the reliability of our benchmark and supports robust evaluation of the VLM multi-agent system's performance.

\textbf{Scene-Level Evaluation Metrics.}
In the scene detection task, we define a true positive (TP) case as correctly predicted as present, a false positive (FP) case as incorrectly predicted as present, and a false negative (FN) case as a missed prediction. Precision is the number of correctly predicted positive cases, and recall is the number of actual positives found.
\textbf{Macro F1.} Macro F1 is the arithmetic mean of per-class F1 scores, prioritizing balanced performance across rare and frequent scene categories\cite{van2004geometry}.
\textbf{Hamming Loss.} Hamming loss normalized misclassification rate that penalizes both FP and FN errors proportionally. A score of 0 indicates perfect scene detection\cite{cesa2004incremental}.

\textbf{Action-Level Evaluation Metrics.}
\textbf{Hierarchical Hamming Loss.} Hierarchical hamming loss penalizes action prediction errors proportionally to scene detection success. This metric is critical for educational systems where scene misidentification (e.g., confusing "group document" with "web content") compounds downstream analysis errors\cite{cesa2004incremental}.
\textbf{Micro-F1.} Micro F1 Aggregates performance across all action instances, favoring models that perform well on frequent behaviors\cite{yang1999re}.

\section{Results}
All evaluations use the 507 held-out test segments, Table \ref{tab:rq2-mas-table} present progressive performance improvements from few-shot baselines (RQ1) to multi-agent enhancements (RQ2).

\subsection{RQ1: Few-shot VLM Performance}
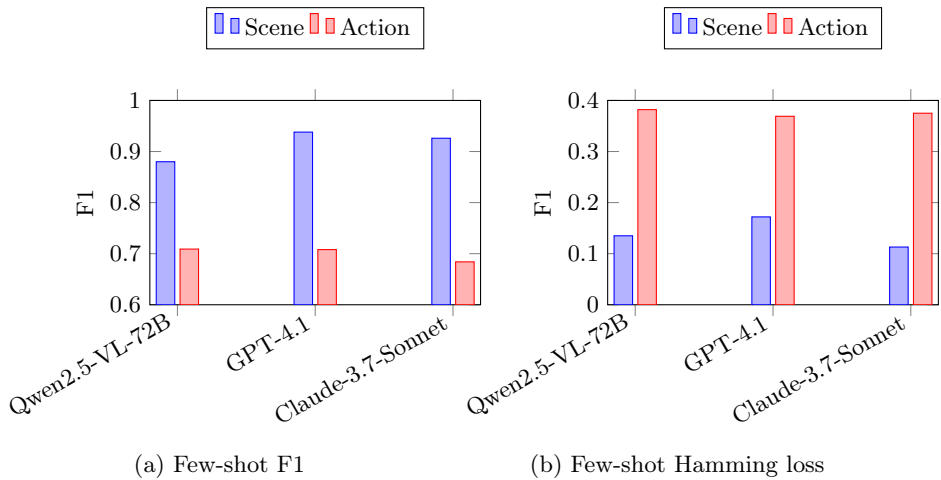
\begin{figure}[t]
\centering
\begin{subfigure}{0.49\textwidth}
\centering
\begin{tikzpicture}
\begin{axis}[
  ybar, bar width=7pt,
  ymin=0.6, ymax=1.0,
  ylabel={F1},
  symbolic x coords={qwen,gpt,claude},
  xtick=data,
  xticklabels={Qwen2.5-VL-72B,GPT-4.1,Claude-3.7-Sonnet},
  xticklabel style={rotate=30, anchor=east},
  legend style={at={(0.5,1.25)}, anchor=south, legend columns=2},
  width=\textwidth, height=4.3cm
]
\addplot coordinates {(qwen,0.880) (gpt,0.938) (claude,0.926)};
\addplot coordinates {(qwen,0.709) (gpt,0.708) (claude,0.684)};
\legend{Scene,Action}
\end{axis}
\end{tikzpicture}
\caption{Few-shot F1}
\end{subfigure}
\begin{subfigure}{0.49\textwidth}
\centering
\begin{tikzpicture}
\begin{axis}[
  ybar, bar width=7pt,
  ymin=0, ymax=0.4,
  ylabel={F1},
  symbolic x coords={qwen,gpt,claude},
  xtick=data,
  xticklabels={Qwen2.5-VL-72B,GPT-4.1,Claude-3.7-Sonnet},
  xticklabel style={rotate=30, anchor=east},
  legend style={at={(0.5,1.25)}, anchor=south, legend columns=2},
  width=\textwidth, height=4.3cm
]
\addplot coordinates {(qwen,0.135) (gpt,0.172) (claude,0.113)};
\addplot coordinates {(qwen,0.382) (gpt,0.369) (claude,0.375)};
\legend{Scene,Action}
\end{axis}
\end{tikzpicture}
\caption{Few-shot Hamming loss}
\end{subfigure}
\caption{RQ1: Few-shot single-VLM performance on scene vs. action detection.}
\label{fig:rq1-single}
\end{figure}

\subsubsection{\textbf{Scene Detection}}
Figure~\ref{fig:rq1-single} shows that single VLMs already perform reasonably on static scene detection, but their reliability varies across architectures. 
GPT-4.1 achieves the highest scene F1 (0.938), while Claude-3.7-Sonnet yields the lowest scene Hamming loss (0.113), indicating fewer per-label errors despite a slightly lower F1 (0.926). 
In contrast, Qwen2.5-VL-72B shows both lower F1 (0.880) and higher Hamming loss (0.135), suggesting weaker consistency under few-shot prompting.

Overall, the mismatch between “best F1” and “lowest Hamming loss” implies that single-VLM scene detection errors are not only about overall correctness, but also about error distribution across labels.
\subsubsection{\textbf{Action Detection}}
For action detection, single VLMs exhibit a clearer limitation under few-shot prompting. All base models reach similar action F1 (0.684--0.709), but action Hamming loss remains high (0.369--0.382), indicating frequent label-wise mistakes in this multi-label dynamic setting. 
Notably, GPT-4.1 attains the lowest action Hamming loss (0.369) while not improving action F1 over Qwen (0.708 vs. 0.709), suggesting that F1 alone can understate differences in per-label error rates.

Compared with scene detection, the consistently larger Hamming losses for actions support that temporal grounding and fine-grained behavior distinction remain challenging for single VLMs in few-shot settings.
\subsection{RQ2: Workflow-based vs. ReAct-style MAS Performance}
\begin{table}[t]
\centering
\caption{RQ2: MAS vs. single VLM performance (F1 / Hamming loss).}
\label{tab:rq2-mas-table}
\resizebox{\textwidth}{!}{
\begin{tabular}{l|cc|cc|cc}
\toprule
\multirow{2}{*}{Model} &
\multicolumn{2}{c|}{Single VLM (RQ1)} &
\multicolumn{2}{c|}{Workflow MAS (RQ2)} &
\multicolumn{2}{c}{ReAct MAS (RQ2)} \\
\cmidrule(lr){2-7}
& Scene (F1/HL) & Action (F1/HL) & Scene (F1/HL) & Action (F1/HL) & Scene (F1/HL) & Action (F1/HL) \\
\midrule
Qwen2.5-VL-72B & 0.880 / 0.135 & 0.709 / 0.382 & 0.965 / 0.038 & 0.763 / 0.316 & 0.951 / 0.046 & 0.771 / 0.287 \\
GPT-4.1        & 0.938 / 0.172 & 0.708 / 0.369 & 0.965 / 0.039 & 0.729 / 0.360 & \underline{0.969} / 0.041 & 0.723 / 0.324 \\
Claude-3.7     & 0.926 / 0.113 & 0.684 / 0.375 & \textbf{0.975} / 0.043 & \underline{0.782} / 0.303 & 0.967 / 0.049 & \textbf{0.793} / 0.263 \\
\bottomrule
\end{tabular}
}
\end{table}

\subsubsection{MAS vs. Single VLM}
Table~\ref{tab:rq2-mas-table} shows that both MAS variants consistently outperform single VLMs on both metrics. 
For scene detection, MAS increases F1 to 0.951--0.975 while reducing Hamming loss to 0.038--0.049, indicating fewer per-scene label errors and more stable predictions. 
For action detection, MAS yields higher action F1 (0.723--0.793) and notably lower Hamming loss (0.263--0.360), demonstrating clearer gains on the error rate of multi-label action assignments.

\subsubsection{Workflow vs. ReAct: Architectural Trade-offs}
Comparing the two MAS designs in Table~\ref{tab:rq2-mas-table}, workflow-based orchestration tends to favor scene understanding: Qwen-workflow and Claude-workflow achieve higher scene F1 than their ReAct counterparts (0.965 vs. 0.951; 0.975 vs. 0.967). 
In contrast, ReAct-style systems generally improve action-level performance: for Qwen and Claude, ReAct yields higher action F1 and lower action Hamming loss than workflow (e.g., Claude: 0.793 / 0.263 vs. 0.782 / 0.303). 
GPT-4.1 shows a smaller and mixed pattern (ReAct slightly higher scene F1, but workflow slightly higher action F1 and higher action HL), suggesting architecture compatibility depends on the base model and task emphasis.

\section{Discussion}
\subsection{Few-Shot VLM Capabilities and Limitations}
Our results achieved a maximum accuracy of 70.55\% in scene-action classification, representing meaningful progress in automated multimodal learning analytics. However, two limitations emerged: \textit{First, prompt sensitivity:} Few-shot VLMs demonstrate strong capabilities in specific, well-defined scenarios. Context-rich prompts yielded 70.55\% accuracy versus lower than 50\% for simple prompts during exploratory studies, suggesting pattern matching guided by linguistic cues rather than robust visual-behavioral understanding. \textit{Second, subtle action detection:} the system exhibited persistent formatting errors and occasional hallucinations, particularly when detecting subtle or ambiguous actions. For example, distinguishing between ``Reading with Highlight'' and ``Reading with Scrolling'' when both involve rapid cursor movement proved challenging, with the model sometimes inventing behavioral details not present in the visual evidence. 

\subsection{Workflow-Based vs. ReAct-Style MAS: Architectural Distinctions and Application Contexts}
Workflow-based MAS emphasizes structured orchestration, decomposing video analysis into predefined subtasks executed in a controlled flow. 
Such agent workflows are often adopted to improve controllability, debuggability, and predictable execution in complex agent systems \cite{yu2025survey}. 
This design aligns with our observation that workflow MAS tends to yield stronger scene-level stability (higher scene F1, lower Hamming loss), where explicit decomposition and constraints help reduce inconsistent scene labeling.

ReAct-style MAS follows an iterative reason-act loop, interleaving explicit reasoning traces with actions to update plans and handle exceptions \cite{yao2022react}.  
This mechanism is compatible with our action detection findings: iterative refinement helps resolve ambiguous evidence and reduces per-label action errors (lower action Hamming loss), especially for fine-grained multi-label predictions. 
In practice, workflow MAS may be preferred when scenes are well-demarcated and efficiency/predictability is critical, whereas ReAct-style MAS is better suited when evidence is ambiguous and iterative verification is necessary. In addition, we need to admit that multi-agent gain and extra-signal gain may be coupled, so our future work will involve ablation experiments to explore the roles of different modules and implement a more robust multi-agent onscreen behavior recognition framework.

\section{Conclusion}
This study contributes to the growing multimodal learning analytics field of AI for education by (1) providing practical evidence on the current capabilities and limitations of VLM-based behavioral analysis; (2) offering a framework for understanding when different agent architectures are most appropriate; and (3) demonstrating that meaningful advances in educational learning analytics require integration of pedagogical theory, behavioral science, and AI methodology. Together, these insights can guide the responsible design of video-analysis systems in education, clarifying when a single VLM is sufficient and when a multi-agent system is necessary to achieve reliable, low-error scene and action classification.

%
%
%
\bibliographystyle{splncs04}
\bibliography{mybibliography}
\end{document}